# Multistage Pruning of CNN Based ECG Classifiers for Edge Devices

Li Xiaolin[1], *Student Member, IEEE*, Rajesh C. Panicker[2] *Member, IEEE*, Barry Cardiff[1] *Senior Member, IEEE*, and Deepu John[1] *Senior Member, IEEE*

*Abstract*— Using smart wearable devices to monitor patients' electrocardiogram (ECG) for real-time detection of arrhythmias can significantly improve healthcare outcomes. Convolutional neural network (CNN) based deep learning has been used successfully to detect anomalous beats in ECG. However, the computational complexity of existing CNN models prohibits them from being implemented in low-powered edge devices. Usually, such models are complex with lots of model parameters which results in large number of computations, memory, and power usage in edge devices. Network pruning techniques can reduce model complexity at the expense of performance in CNN models. This paper presents a novel *multistage pruning* technique that reduces CNN model complexity with negligible loss in performance compared to existing pruning techniques. An existing CNN model for ECG classification is used as a baseline reference. At 60% sparsity, the proposed technique achieves 97.7% accuracy and an F1 score of 93.59% for ECG classification tasks. This is an improvement of 3.3% and 9% for accuracy and F1 Score respectively, compared to traditional *pruning with fine-tuning* approach. Compared to the baseline model, we also achieve a 60.4% decrease in run-time complexity.

## I. INTRODUCTION

This work is aimed at optimizing the complexity of ECG classification CNNs using network pruning techniques. Generally, deep learning model has lots of parameters, which require huge amount of storage and computational complexity, and consequently, time and energy. This makes it difficult to deploy deep neural networks on wearable devices. Pruning a network can reduce the number of parameters to achieve the aim, and hence the complexity of neural networks [1]. Based on the structure of sparsity, network pruning can be classified into fine-grained sparsity, vector level sparsity, kernel level sparsity, and filter level sparsity [2]. Filter-wise and channel-wise pruning have a higher possibility of cutting off key neurons, thus degrading the performance of the model. Therefore, we use weight-wise pruning (fine-grained sparsity) to implement network pruning optimisation. In this paper, we propose a novel method to prune a 1-dimensional CNN to classify ECG signal with minimal loss of accuracy. Further, we compare it with other algorithms.

Section II illustrates some background work related to network pruning algorithms. We introduce our baseline model in Section III. Section IV shows the three algorithms we proposed; the results obtained and discussion are in Section V. Section VI concludes our work on network pruning.

*This work was supported in part by the China Scholarship Council, the Microelectronic Circuits Centre Ireland, and the Irish Research Council under the New Foundations Scheme.
[1]University College Dublin, xiaolin.li@ucdconnect.ie, {barry.cardiff, deepu.john}@ucd.ie
[2]National University of Singapore, rajesh@nus.edu.sg

## II. RELATED WORK

Convolutional neural network (CNN) is a powerful machine learning tool, which is increasingly being used in practical applications [3]–[5]. However, the sizes of CNN models are typically too big to be deployed in wearable devices. Hence, there has been a lot of interest in compressing CNNs. Song *et al.* compressed the network by pruning away unimportant connections and reduced the number of weights by a factor of ten [6]. Yang *et al.* proposed a soft filter pruning approach that can reduce more than 42% FLOPs on ResNet-101 [7]. Zhuang *et al.* proposed a channel-wise sparsity pruning method in the optimization objective during training [8]. They justify their work only on public images dataset using network pruning on different levels. We propose a new method of pruning for ECG classification in this paper.

## III. BASELINE MODEL

Fig. 1 illustrates one-dimensional convolutional neural network for ECG classification from single-lead ECG [9]. We use this as our baseline model. At first, we extract QRS complex from the dataset based on annotations. The ECG signals were sampled at 360 Hz and each QRS complex has 260 samples [10]. The input of this network is a single QRS complex. Thus, each sample has $\frac{1}{360}$ seconds, and the length of each input heartbeat is around 0.72 seconds. The output will be one of five specific heartbeat types based on AAMI standard [11]. The model can classify 5 different heartbeat types, N, SVEB, VEB, F, and Q, through these 10 layers.

This model achieves 98.12% overall accuracy with 98.07% sensitivity and 98.29% specificity. Table I displays the accuracy, sensitivity, F1 Score, etc. Based on the results, this model can classify ECG into N, SVEB, VEB, F, Q correctly using MIT-BIH Arrhythmia Database [12].

Table I shows the performance metrics for individual classes and the overall dataset, balanced using SMOTE algorithm [13]. It is slightly worse compared to the original imbalanced dataset. However, these results reflect a more real-world performance of the proposed technique.

## IV. PRUNING METHODS

Considering the baseline model, Fig. 1, the majority of the trained parameters, i.e., the weights and the biases,

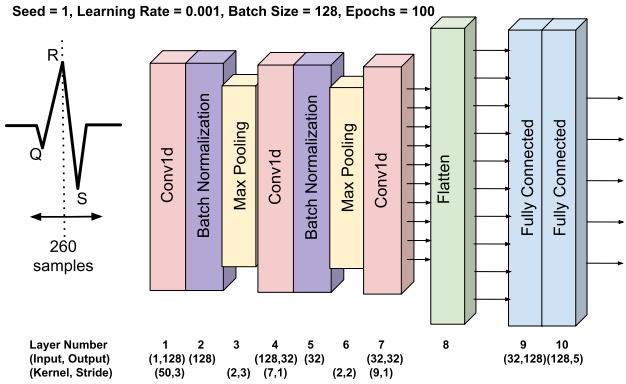

Fig. 1: Baseline CNN architecture for classification [9].

TABLE I: Metrics of the baseline model

|  | N | SVEB | VEB | F | Q | Total |
|---|---|---|---|---|---|---|
| Accuracy | 98.37% | 98.92% | 99.63% | 99.65% | 99.91% | 98.12% |
| Sensitivity | 98.37% | 93.60% | 98.03% | 88.00% | 99.60% | 98.07% |
| Specificity | 98.36% | 99.06% | 99.75% | 99.74% | 99.93% | 98.29% |
| Precision | 99.65% | 72.48% | 96.41% | 71.90% | 99.20% | 92.33% |
| F1 Score | 99.01% | 81.70% | 97.21% | 79.14% | 99.40% | 95.04% |

are contained in the three convolutional layers. Of these, the weights involve multiplication and hence add more to the computational complexity, whereas each bias only contributes a single addition in each neuron. Thus, in this work, we focus on complexity reduction by eliminating (or *pruning*) as many weights as possible so as not to impact adversely on the performance. Fig. 2 shows the process of pruning connections between neurons within each of these convolutional layers.

We proposed three magnitude-based network pruning algorithms. All of the pruning algorithms presented below use the baseline model as a starting point. The pruning process involves three steps - training a large model, followed by pruning weights, and eventually, fine-tuning [14] the weights. Here we consider that the initial training (i.e., the baseline model) has already been done [9], and we focus on the pruning and fine-tuning which are presented in the following sections.

### A. Simple pruning

The obvious method to prune the neural network is to simply prune $\eta$ the weights based on their magnitude, with $\eta$ ranging from 0 to 1, usually expressed as a percentage (0% to 100%). To do this, we rank the individual weights for the three layers separately and set the smallest $\eta\%$ of them to zero. In effect, this cuts off the connections between those neurons. Fig. 3a displays the process of the simple pruning

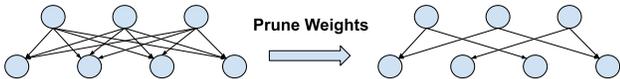

Fig. 2: Pruning weights (connections) in convolutional layers.

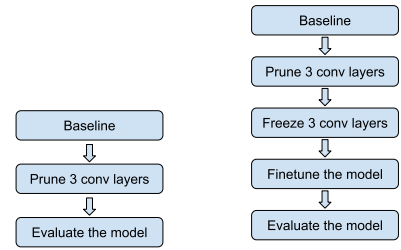

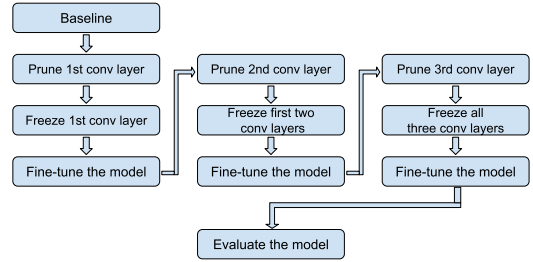

Fig. 3: Three magnitude-based network pruning algorithms: (a) simple pruning, (b) pruning with fine-tuning, (c) multistage pruning.

method, and is referred to in the results section V as *simple pruning*.

### B. Pruning with fine-tuning

Here we consider an improvement in the above *simple pruning* whereby, after zeroing, we fine-tune all the other parameters in the baseline model, i.e., everything except the three convolutional layers. In effect, we are firstly cutting off the connection between the neurons based on the initial training and then re-training. Fig. 3b shows the process of pruning with fine-tuning method.

### C. Multistage pruning

Rather than performing the pruning / fine-tuning on all three convolutional layers separately, here we present a novel variation whereby we perform the zeroing and fine-tuning on each layer in order.

We start by ranking the weights for the first layer and zeroing (and freezing) the required $\eta\%$ of these before re-training / fine-tuning the remaining parameters in the baseline model. We then progress onto the second convolutional layer and zero the required percentage of those and perform the fine-tuning. Finally, we repeat this process for the third convolutional layer. Fig. 3c displays the process of multistage pruning method.

## V. RESULTS & DISCUSSION

We apply the three pruning strategies on the baseline model using MIT-BIH Arrhythmia dataset. For each method, we tried different sparsity levels ($\eta$), ranging from 10% to 90%. According to the specified sparsity ($\eta$), different thresholds will be set to achieve removing different percentages of connections in the model. For each sparsity ($\eta$) level, we evaluate the performance as well as the complexity in the following sub-sections.

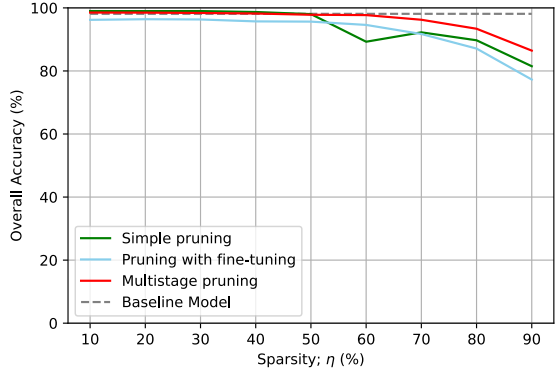

Fig. 4: Accuracy analysis.

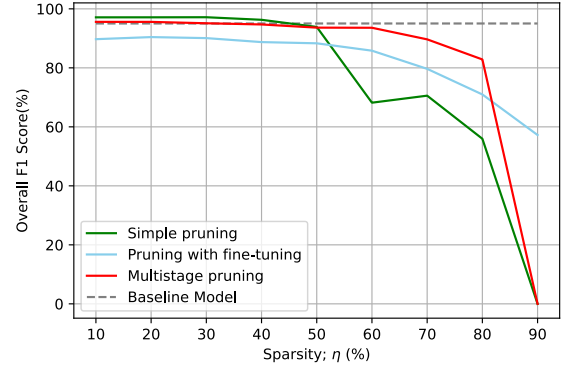

Fig. 5: F1 score analysis.

## A. Performance versus Sparsity

To evaluate the model, 70%, 15%, 15% of the data is used for fine-tuning, validation, and testing respectively.

Various figures of merits used are as given below:

$$\text{Accuracy} = \frac{TN + TP}{TN + TP + FP + FN} \quad (1)$$

$$\text{Sensitivity (Recall)} = \frac{TP}{TP + FN} \quad (2)$$

$$\text{Precision} = \frac{TP}{TP + FP} \quad (3)$$

$$\text{F1 Score} = \frac{2 \times \text{Recall} \times \text{Precision}}{\text{Recall} + \text{Precision}} \quad (4)$$

where,
- TN = True Negative, the number of normal beats correctly classified as being normal.
- FN = False Negative, the number of normal beats falsely classified as normal.
- TP = True Positive, the number of non-normal (i.e. S, V, F, and Q) beats correctly classified.
- FP = False Positive, the number of non-normal beats incorrectly classified.

For each of the three algorithms (and the baseline model), we measure the overall accuracy, F1 score, loss, and sensitivity for the different levels of sparsity; these are calculated in Equations. (1), (2), (3), (4) separately, and plotted in Figures. 4, 5, 6 and 7 respectively.

Fig. 4 shows the overall accuracy as a function of sparsity level. It can clearly be seen that pruning has a large impact. It can be seen that the *multistage pruning* offers the best performance of the three algorithms by maintaining good accuracy up to a sparsity level of c.60% to 70%, whereas *pruning with fine-tuning* is only usable up to a sparsity level of c.50%. The *simple pruning* method fails early on and is clearly not a good choice for accuracy.

From the F1 score in Fig. 5, we see that the *multistage pruning* offers the best performance of the three algorithms with very little degradation up to c.60% to 70%. The same trend is apparent in Fig. 6 for the overall loss.

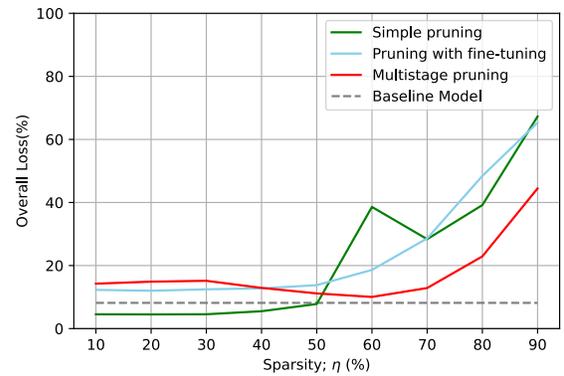

Fig. 6: Loss analysis.

Fig. 7 shows results of sensitivity as a function of the sparsity level. The sensitivity decreases rapidly after c.30% of the weights are pruned when employing the *simple pruning*. However, it is noteworthy to see that the sensitivity is largely unaffected for the other two algorithms, underscoring the need for the fine-tuning step(s).

## B. Complexity versus Sparsity

Considering that our purpose is to reduce the run-time power consumption, we ignore the complexities associated with the training procedure and focus instead on the complexity of the pruned neural network. The number of FLOPs in each layer as a function of $\eta$ is shown in Table III. Fig. 8 shows the overall trend of FLOPs with sparsity $\eta$. It can be observed that FLOPs required decreases with $\eta$. With 60% pruning, we can observe that the complexity is reduced from 1.01 million FLOPs in the baseline model to 0.4 million FLOPs in the pruned model; which corresponds to a 60.4% complexity reduction, which can result in significant power savings. This is enumerated in Table II, along with the associated algorithm metrics. At 60% sparsity, the multistage pruning technique achieves 97.7% accuracy and an F1 Score of 93.59%. This is an improvement of 3.3% and

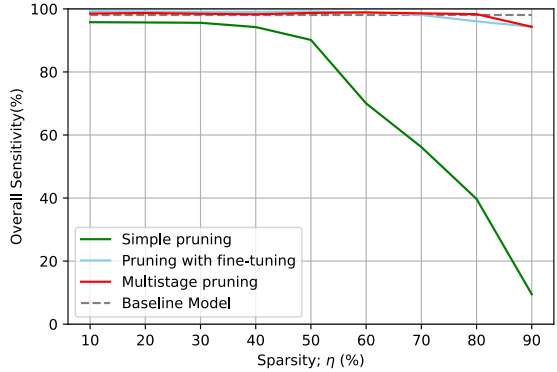

Fig. 7: Sensitivity analysis.

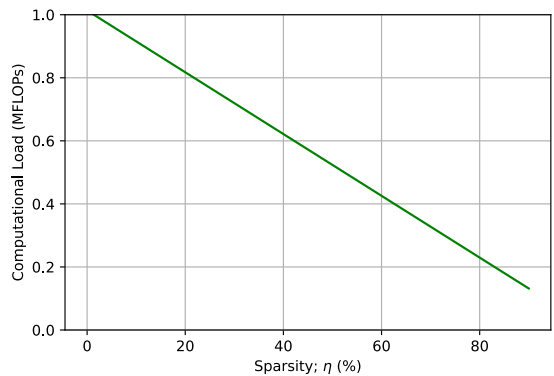

Fig. 8: FLOPs analysis.

9% compared to *pruning with fine-tuning* approach.

TABLE II: Comparison of baseline and pruned models.

|  | Accuracy | Sensitivity | F1 Score | FLOPs |
|---|---|---|---|---|
| baseline model | 98.12% | 98.07% | 95.04% | 1013472 |
| 60% *pruning with fine-tuning* | 94.60% | 98.86% | 85.82% | 425645 |
| 60% *multistage pruning* | 97.72% | 98.88% | 93.59% | 425645 |

TABLE III: Floating point operations (FLOPs) for each layer.

| Layer | # FLOPs |
|---|---|
| 1 | $71 \times 50 \times (1-\eta) \times 128 \times 2$ |
| 2 | $71 \times 128$ |
| 3 | 0 |
| 4 | $18 \times 7 \times (1-\eta) \times 32 \times 2$ |
| 5 | $18 \times 32$ |
| 6 | 0 |
| 7 | $1 \times 9 \times (1-\eta) \times 32 \times 2$ |
| 8 | 0 |
| 9 | $32 \times 128 \times 2$ |
| 10 | $128 \times 5 \times 2$ |
| Sum | $917440 \times (1-\eta) + 19136$ |

## VI. CONCLUSIONS

In this work, we proposed and compared the performance of three pruning algorithms for use with CNNs for low-power ECG classification. The first, *simple pruning*, just prunes all convolutional layers at once without any retraining. The next, *pruning with fine-tuning*, does pruning followed by retraining of the remaining weights. The third, *multistage pruning*, is a novel stage-by-stage pruning / retraining algorithm. For all three cases, the run-time complexity of the resulting neural networks is identical for a given pruning level, but their performance differs greatly. The performance of the three algorithms were evaluated through extensive simulations.

Sensitivity, a vital metric in biomedical applications, of *simple pruning* is shown to degrade very quickly with sparsity level and is thus not a viable method. However, the sensitivity of the other two algorithms remains robust even at high sparsity levels. *Multistage pruning* has superior performance in all metrics especially at high levels of pruning as compared to *pruning with fine-tuning*. When compared to the baseline model, we find that the *multistage pruning* model has almost no performance degradation for all pruning levels not exceeding 60%. Future work includes a more extensive evaluation of the proposed pruning strategies on other datasets as well as other CNN architectures.